\documentclass[10pt,twocolumn,letterpaper]{article}

\usepackage[pagenumbers]{wacv} 

\usepackage{graphicx}
\usepackage{amsmath}
\usepackage{amssymb}
\usepackage{booktabs}

\usepackage{url}

\usepackage{epsfig}
\usepackage{makecell}
\usepackage{xcolor}
\usepackage{float}
\usepackage{subcaption}
\usepackage{wrapfig,lipsum}
\usepackage{adjustbox}

\usepackage{soul}

\usepackage[accsupp]{axessibility}

\newtheorem{theorem}{Theorem}
\newtheorem{proposition}[theorem]{Proposition}

\newcommand{\network}{AHFNet} 

\newcommand{\module}{high-frequency extraction module}

\newcommand{\h}{0}
\newcommand{\wa}{0.15}
\newlength \g
\newcommand{\name}{0}
\newcommand{\namepatch}{0}
\newcommand{\w}{0.15}

\usepackage[pagebackref,breaklinks,colorlinks]{hyperref}

\usepackage[capitalize]{cleveref}
\crefname{section}{Sec.}{Secs.}
\Crefname{section}{Section}{Sections}
\Crefname{table}{Table}{Tables}
\crefname{table}{Tab.}{Tabs.}

\def\wacvPaperID{505} 
\def\confName{WACV}
\def\confYear{2025}

\begin{document}

\title{Adaptive High-Pass Kernel Prediction for Efficient Video Deblurring}

\author{Bo Ji \qquad Angela Yao \\
National University of Singapore\\
{\tt\small \{jibo,ayao\}@comp.nus.edu.sg}
}

\maketitle

\begin{abstract}
State-of-the-art video deblurring methods use deep network architectures to recover sharpened video frames. Blurring especially degrades high-frequency (HF) information, yet this aspect is often overlooked by recent models that focus more on enhancing architectural design. Recovering these fine details is challenging, partly due to the spectral bias of neural networks, which are inclined towards learning low-frequency functions. To address this, we enforce explicit network structures to capture the fine details and edges. We dynamically predict adaptive high-pass kernels from a linear combination of high-pass basis kernels to extract high-frequency features. This strategy is highly efficient, resulting in low-memory footprints for training and fast run times for inference, all while achieving state-of-the-art when compared to low-budget models. The code is available at \href{https://github.com/jibo27/AHFNet}{https://github.com/jibo27/AHFNet}.
\end{abstract}

\section{Introduction}\label{sec:introduction}

Video deblurring sharpens video frames from blurry input sequences.  
The blurring process suppresses the high frequencies in the original video, so deblurring can be viewed as restoring the lost frequencies.
Classical sharpening techniques enhance the edges and detailing using unsharp masking and inverse filtering~\cite{ye2018blurriness,deng2010generalized}.  Others~\cite{xu2013unnatural,krishnan2011blind,pan2016l_0}
try to estimate the underlying blur kernels to reverse the effects. In the case of multi-image deblurring~\cite{zhang2013multi,cai2009blind}, 
the missing information can be approximated by observations over multiple frames. 
Classical methods are often interpretable and based on well-understood mathematical principles.
However, these classical methods are usually based on degradation models that are too simple for real-world scenes.

Recent advances in video deblurring rely on deep neural networks trained end-to-end, achieving state-of-the-art performance through local alignment~\cite{wang2022efficient,pan2020cascaded,lin2022flow} and deformable convolutions~\cite{wang2019edvr,jiang2022erdn}.  While effective, these techniques are computationally expensive and non-ideal for optimized hardware.
Specifically, the memory footprint of such models has exploded, 
requiring over 200GB of GPU memory~\cite{lin2022flow,liang2022recurrent,li2023simple,Pan_2023_CVPR} to process frames of size 
$256\times 256$ in a batch.
This surge in the training budget not only incurs high financial costs but also leads to significant environmental impact~\cite{bartoldson2023compute}. 
A larger training budget often indicates a complicated model, which requires more inference time. 
Apart from the efficiency issues, deep learning-based models~\cite{li2023simple,liang2022recurrent} have made mostly black-box gains compared to classical deblurring methods.
They are relatively agnostic to the underlying deblurring problem, even though deblurring is a well-studied problem in image processing~\cite{gonzales1987digital,ye2018blurriness}. 
The interpretability of the functionality of the network modules remains obscure.

Moreover, standard neural network architecture components are counterproductive for deblurring as they tend to suppress high-frequency features, which are crucial for video deblurring.  
For example, self-attention in transformers acts as a low-pass filter~\cite{park2022vision,wang2022anti}; convolutional networks also tend to weaken the high-frequency components~\cite{tang2022defects}.
Finally, neural networks are prone to spectral bias - a preference for learning lower-frequency components \cite{rahaman2019spectral}.

To address these issues, we propose an \textbf{A}daptive \textbf{H}igh-\textbf{F}requency extraction network (\textbf{\network}) for video deblurring.  
Our method maintains interpretability by explicitly leveraging high frequencies extracted from blurry videos for deblurring. 
To ensure effective extraction, we define a set of high-pass basis kernels. 
A high-pass kernel is a kernel that emphasizes features like edges and local details.
Then, we dynamically predict mixing coefficients to perform linear combinations on the basis kernels, resulting a new high-pass kernel.
The new generated kernels are further convolved with the blurry features to produce crucial details required for deblurring.
Our experiments demonstrate that incorporating even the simplest high-pass kernels, such as Sobel filters, can significantly improve deblurring performance. 
For instance, integrating spatial and temporal gradients into a neural network for video deblurring results in a notable improvement of 0.33 dB (see Section~\ref{sec:choice_building_kernels}). 

Moreover, our strategy is lightweight and efficient in both training and inference, as our building blocks require only a few convolutional layers. The memory footprint for training our model is only one-sixth of that required for state-of-the-art models (see Fig.~\ref{fig:tradeoff_psnr_training_cost}). 
Compared to other models with similar low-budget training requirements, our method achieves competitive performance on standard deblurring benchmarks~\cite{nah2017deep,su2017deep}.
For instance, our model achieves an inference speedup of 35x compared to ERDN{~\cite{jiang2022erdn}} while achieving superior PSNR and SSIM scores.
(See Fig.~\ref{fig:teaser_photo} and Table~\ref{table:sota_gopro}).
Our contributions can be summarized as follows:
\begin{itemize}
    \item We highlight the critical role of high frequencies in video deblurring and reformulate the deblurring     task based on explicitly extracting and applying high-frequency information. 
    \item We introduce a new \module{} to adaptively recover spatial and temporal detailing for video deblurring.
    This module forms our proposed \network{} for efficient yet effective video deblurring.  
    \item Our proposed \network{} delivers top accuracy performance among models under similar memory footprints while being more efficient in inference time and GMACs.  
\end{itemize}

\section{Related Work}

\noindent
\textbf{Classic deblurring.} Classic deblurring algorithms often leverage image structures or priors, \eg , sparse gradient priors~\cite{xu2013unnatural,krishnan2011blind}, intensity priors~\cite{pan2016l_0}, and edges~\cite{cho2009fast,xu2010two,yang2019variational,sun2013edge,xu2013unnatural}. 
Unsharp masking~\cite{ye2018blurriness,deng2010generalized,polesel2000image} counteracts blurring in an image by reintroducing a scaled version of the image's high-frequency components. This emphasizes the fine details or edges 
and enhances the image's overall sharpness.
Other methods iteratively estimate the blur kernel and sharpened outputs~\cite{cho2009fast,zhang2022pixel,pan2016blind,shan2008high}. Multi-image blind deconvolution algorithms restore the target image from multiple blurred or relevant images~\cite{cai2009blind,rav2005two,zhu2012deconvolving,zhang2013multi,he2012guided}. 
Classic deblurring algorithms have difficulty generalizing well to diverse and unseen types of blur.
Our work explicitly utilizes high-frequency information with the substantial capacity of neural networks to enhance video deblurring. 

\noindent
\textbf{Deep learning methods.} Deep learning methods 
restore sharp videos with deep neural networks~\cite{su2017deep,wang2019edvr,pan2020cascaded,zhong2020efficient,ji2022xydeblur}. 
XYDeblur~\cite{ji2022xydeblur} introduces an one-encoder-two-decoder by rotating decoders to predict complementary deblurring information. Though both XYDeblur and our approach use rotation, XYDeblur rotates decoder parameters for image deblurring; we rotate fixed kernels to enhance frequency for video deblurring.
Recent works have focused on 
information retrieval from neighboring frames with optical flow~\cite{wang2022efficient,lin2022flow}, deformable convolution~\cite{wang2019edvr,jiang2022erdn}, and feature matching~\cite{ji2022multi,li2021arvo}. 
RVRT~\cite{liang2022recurrent} combines optical flow, deformable convolution, and attention mechanisms to achieve high restoration quality. ShiftNet~\cite{li2023simple} uses spatial-temporal shifts to enhance video aggregation.
Yet these operations are computationally expensive and not well-suited for standard hardware. For example, deformable convolution requires increased memory access, irregular data structuring, and dynamic computational graphs, all adding to the complexity and resource demands~\cite{guan2022memory}. 
Rather than focus on the architecture design, our work takes a distinct direction by placing emphasis on the significance of high frequencies to improve deblurring outcomes. 

\noindent
\textbf{Kernel prediction networks.} Kernel prediction networks determine convolution weights dynamically during inference~\cite{chen2020dynamic,ma2020weightnet,zhou2019spatio,jia2016dynamic}. 
The kernel prediction network~\cite{xia2020basis,zhang2023kbnet} synthesizes kernel bases and coefficients to derive adaptive filters. 
The optimization of kernel prediction modules is challenging and requires extensive training data and extended training time 
to accurately predict a kernel. 
Our method incorporates specific basis kernels designed to emphasize the extraction of high-frequency components. We solely predict coefficients for their linear combination, which simplifies the computational load.

\section{Approach}\label{sec:approach}

\subsection{Preliminaries}

The video blurring process can be viewed as a low-pass filtering on consecutive latent sharp frames~\cite{nah2017deep,su2017deep}.
We follow the video blur accumulation model from~\cite{nah2017deep}, which forms a blurry frame, $x_t$, by accumulating sampled sharp frames $s_{t,i}$:
\begin{equation}\label{eq:blur_temporal}
    x_t \approx g\left(\frac{1}{B}\sum_{i=1}^{B}s_{t,i}\right), 
\end{equation}
where $g$ is the nonlinear camera response function, $B$ is the number of accumulated frames and $s_{t,i}$ is the original captured frame related to $x_t$. 
In the simplest setting where the camera response function is assumed to be an identity 
function, Eq~\ref{eq:blur_temporal} simplifies to a convolution operation of consecutive images with an averaging kernel.
It is well-established that averaging has the effect of low-pass filtering~\cite{jahne2005digital}.
Consequently, a blurry frame can be regarded as the low-pass filtered outcome of the latent sharp frames. Therefore, our main goal is to recover the lost high frequencies.

\begin{figure*}[t!]
    \centering
    \includegraphics[width=0.8\linewidth]{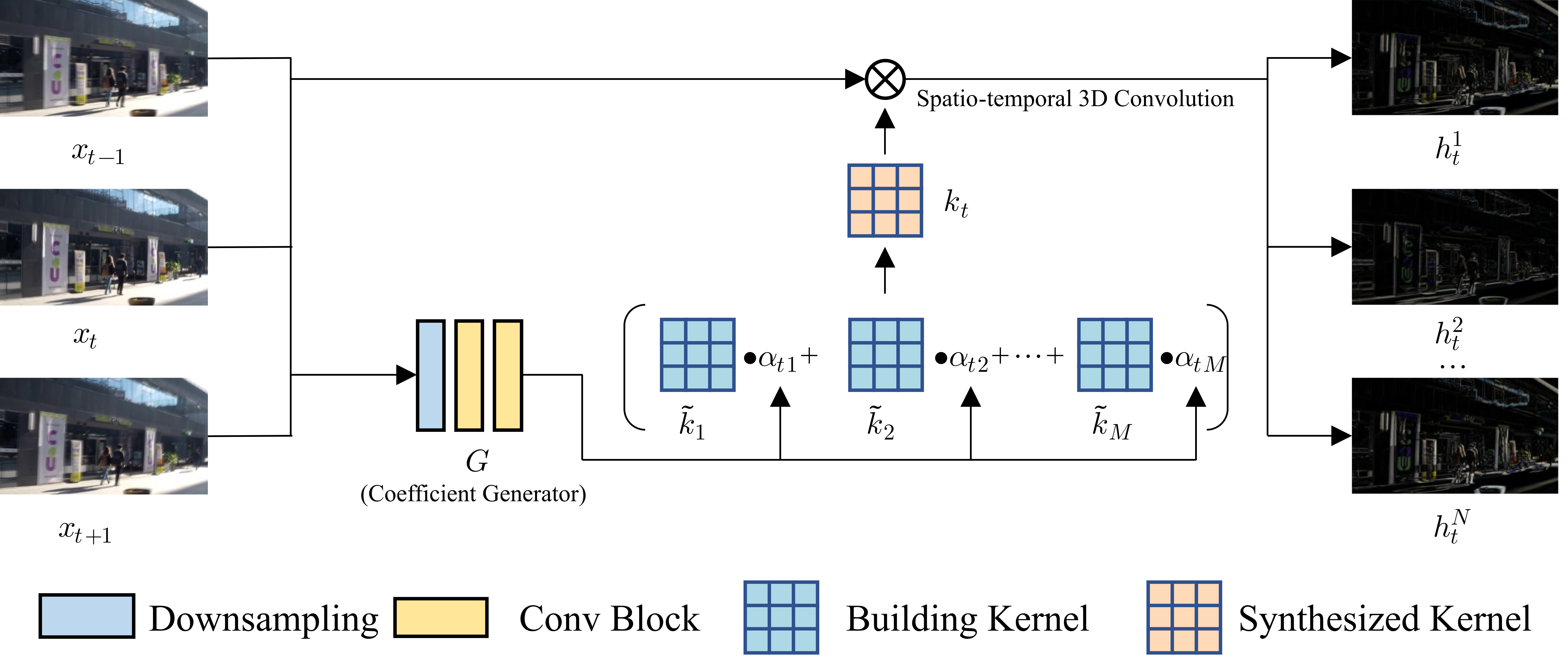}
    \vspace{-0.2cm}
    \caption{
    Adaptive \module{} ($\mathcal{H}$). We generate the dynamic high-pass kernel $k_t$ for future convolution by performing a linear combination of high-pass basis kernels $\{\tilde{k}_j\}_{j=1}^M$ and the coefficients $\{\alpha_{t,j}\}_{j=1}^M$.}\label{fig:module}
    \vspace{-0.5cm}
\end{figure*}

\subsection{Formulation}
Consider a local temporal window of span $l$ around the $t$-th blurry frame $x_t$ in a video, denoted as $\{x_i\}_{i=t-l}^{t+l} = \{x_{t-l}, \dots, x_t, \dots x_{t+l}\}$. For a residual network, the deblurred frame $\hat{y}_t$ can be expressed as: 
\begin{align}
    \hat{y}_t = x_t + \mathcal{L}(x_t, \mathcal{H}(\{x_i\}_{i=t-l}^{t+l})),\label{eq:formulation}, 
\end{align}
\noindent \ie the sum between the original input $x_t$ and some residual component extracted by $\mathcal{L}$. 
Specifically, the residual component depends on both the current input $x_t$ as well as an operation $\mathcal{H}$ applied to 
a temporal window of data $\{x_i\}_{i=t-l}^{t+l}$, since blurring often involves inter-frame effects (see  Eq.~\ref{eq:blur_temporal}). 

In the simplest setting when
$\mathcal{H}$ is set as a simple high-pass filter and $\mathcal{L}$ operates as scalar multiplication independent of the input, Eq~\ref{eq:formulation} reduces 
to the classic unsharp masking used for image sharpening:
\begin{align}
\hat{y}_t = x_t + \lambda (k * x),
\label{eq:unsharp_masking}
\end{align}
where $*$ denotes a 2D convolution,  $\lambda$ is a scaling factor 
and $k$ can be any high-pass filter. 
Eq~\ref{eq:unsharp_masking} shows that unsharp masking explicitly applies high-pass filtering to blurry inputs and adds back the filtered result to get the deblurred output.

Our general formulation in Eq~\ref{eq:formulation} is more flexible and expressive than the standard unsharp masking because we incorporate inter-frame analysis and use a convolutional network to predict the features that play a similar role to $(k*x)$ and $\lambda$. Based on our formulation, we break down the deblurring process into two primary phases: 1) extracting high frequencies from both the current and adjacent frames with $\mathcal{H}$; and 2) converting these features obtained from blurry frames back with $\mathcal{L}$, effectively creating a feature map for reconstructing the latent sharp frame.

\subsection{Adaptive \module}\label{sec:module}

Both transformers and convolutional networks have challenges in extracting high frequencies. The self-attention mechanisms in transformers act as low-pass filters and attenuate high-frequency details in the input features~\cite{wang2022anti,park2022vision}.  Specifically, the low-pass nature of self-attention comes from the softmax operation that applies a non-negative, weighted averaging over the inputs.  As such, previous works~\cite{park2022vision} have found that transformers have stronger low-pass effects than convolutional operations. Yet 
from a statistical perspective, 
feed-forward
convolutional networks are also likely to weaken high-frequency components~\cite{tang2022defects},
as the Fourier coefficients mostly locate in low frequencies after the backward propagation for the convolutional weights initialized from a Gaussian distribution.

To mitigate this, we use convolutional networks while constraining the learning space of the convolutional kernels. 
Our goal is to ensure that the generated features represent the high-frequency components of the input. The strategy is to make the operations similar to high-pass filtering. 
Specifically, we propose a module $\mathcal{H}$, consisting of kernels  which function only as high-pass filters (see Figure~\ref{fig:module}). 
Before delving into the details of $\mathcal{H}$, we establish the following proposition on the linear combination of high-pass filters: 
\begin{proposition}\label{prop:combination_filter}
Consider $M$ spatial high-pass filters with impulse responses $h_i(x)$ and corresponding frequency responses $H_i(f)$ with cutoff frequencies $f_{ci}$, sorted such that $f_{c1} \leq f_{c2} \leq \dots \leq f_{cM}$. Any linear combination of these filters with non-negative coefficients $\alpha_i$ in the spatial domain, represented as
\begin{align}
    h(x) = \sum_{i=1}^M \alpha_i h_i(x),
\end{align}
will also act as a high-pass filter in the spatial domain, with a corresponding frequency response $H(f)$ and a cutoff frequency not greater than $f_{c1}$.
\end{proposition}

Proposition~\ref{prop:combination_filter} states that any non-negative linear combination of high-pass filters remains a high-pass filter.  The proof is given in the supplementary material.  
This proposition motivates us to train a kernel prediction module in which the kernel weights are formed by a linear combination of predetermined high-pass kernels and dynamically predicted coefficients.
Without imposing this high-pass constraint, arbitrarily learned 
convolutional kernels may not consistently optimize for high-pass filtering.  
We verify this empirically in Section~\ref{sec:hf_kernel_operation}.

Now consider a set of 3D high-pass basis kernels $\{\tilde{k}_j\}_{j=1}^M=\{\tilde{k}_1, \dots, \tilde{k}_M\}$. 
Each kernel is of size $\mathbb{R}^{T_k\times H_k\times W_k}$ where $T_k$, $H_k$, and $W_k$ represent the temporal length, height, and width of the kernel, respectively.
The input has $T_k$ consecutive frames of size $C\times T_k\times H\times W$, where $C$ is the input dimension.
First, we generate a dynamic kernel $k_t$ from mixing coefficients $\alpha_t$:
\begin{align}\label{eq:kernel_base_to_kernel}
    k_t = \sum_{j=1}^{M}\alpha_{t,j} \tilde{k}_j, \;\; \text{where} \;\; \{\alpha_{t,j}\}_{j=1}^M = \mathcal{G}\left(\{x_i\}_{i=t-l}^{t+l}\right), 
\end{align}
where $\mathcal{G}$ denotes the coefficient generator and $\alpha_{t,j}$ denotes the coefficient for the $j$-th kernel of $x_t$.  
The kernel $k_t$ has a size of $1\times T_k\times H_k\times W_k$.  Proposition~\ref{prop:combination_filter} assures that $k_t$ remains a high-pass kernel. 
 
\begin{figure*}[t!]
    \centering
    \includegraphics[width=0.8\linewidth]{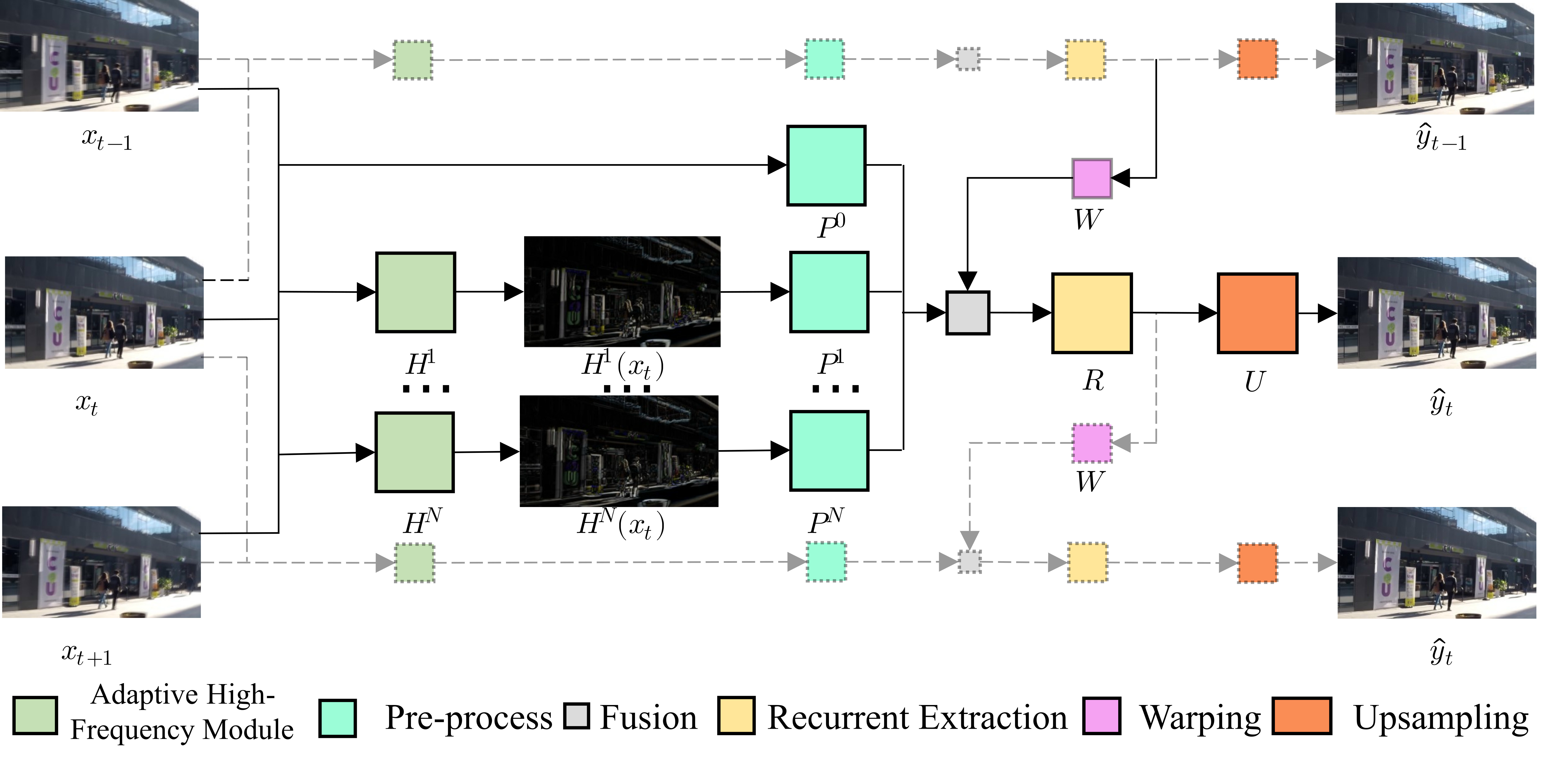}
    \vspace{-0.2cm}
    \caption{Overview of \network{}. We extract high frequencies using $\mathcal{H}$, which are explicitly utilized for deblurring.}\label{fig:network}
    \vspace{-0.5cm}
\end{figure*}

Convolving image $x_t$ in space-time with $k_t$ leads to the high-frequency feature $h_t$. To cover orthogonal directions, we rotate the kernel set $\{\tilde{k}_i\}_{i=1}^M$ by 90 degrees anticlockwise, forming a new set of orthogonal basis kernels, which we denote as 
$\{\tilde{k}_i'\}_{i=1}^M$
.
Next, we apply the coefficients $\{\alpha_{t,j}\}_{j=1}^M$ to the set of rotated basis kernel, akin to the method in Eq~\ref{eq:kernel_base_to_kernel}, resulting in another feature, denoted as $h_t'$.  This process can be summarized as
\begin{align}
    h_t = k_t * \{x_i\}_{i=t-l}^{t+l}; \  h_t' = k_t' * \{x_i\}_{i=t-l}^{t+l}.
\end{align}
For clarity, $\tilde{k}_j$ and $\tilde{k}_j'$ represent the original and rotated basis kernels, respectively. $k_t$ and $k_t'$ denote the dynamically predicted corresponding kernels that are further convolved with the input feature $x_t$.

While we can learn distinct coefficients for the rotated basis kernel $\{\tilde{k}'_i\}_{i=1}^M$, applying the same coefficients achieves comparable performance. Finally, we combine the features extracted in orthogonal directions by summing the absolute values of the feature maps $h_t$ and $h_t'$. The output can be expressed as:
\begin{align}
    \mathcal{H}(\{x_t\}_{i=t-l}^{t+l}) = \{h_t, h_t', |h_t| + |h_t'|\}.
\end{align}
For brevity, we use $\mathcal{H}(x_t)$ to represent $\mathcal{H}(\{x_t\}_{i=t-l}^{t+l})$. Incorporating rotated kernels enhances 
diversity in the extracted frequencies while introducing the sum of absolute values facilitates a more comprehensive information aggregation from the two directions.

Predicting the mixing coefficients $\alpha$ leads to the dynamic generation of convolutional weights tailored to match input characteristics. An alternative approach is dynamic convolution~\cite{jia2016dynamic}, which predicts the final high-pass kernel $k_t$ directly instead of the mixing coefficients $\alpha$. Compared to this alternative, our method reduces the number of learnable weights from $T_kH_kW_k$ to $M$, simplifying the optimization process. Consequently, our approach is more efficient in both training and inference. Moreover, our method ensures that the generated kernel is high-pass (see Proposition~\ref{prop:combination_filter}) and has proven effective (see Section~\ref{sec:hf_kernel_operation}).

In our implementation, 
$\mathcal{H}$
adopts $M\!=\!4$ 
basis kernels. Two are normalized $3 \times 3$ Sobel kernels in the horizontal and vertical direction.
The other twos are temporal differencing kernels $\begin{bmatrix}
\begin{bmatrix}0\end{bmatrix},
\begin{bmatrix}-1\end{bmatrix},
\begin{bmatrix}1\end{bmatrix}
\end{bmatrix}$ and $\begin{bmatrix}
\begin{bmatrix}1\end{bmatrix},
\begin{bmatrix}-1/2\end{bmatrix},
\begin{bmatrix}-1/2\end{bmatrix}
\end{bmatrix}$, denoting $- x_t + x_{t+1}$ for the first and $x_{t-1} - 0.5x_t - 0.5x_{t+1}$ for the second. 
The second temporal kernel is derived from $\begin{bmatrix}
\begin{bmatrix}1\end{bmatrix},
\begin{bmatrix}-1\end{bmatrix},
\begin{bmatrix}0\end{bmatrix}
\end{bmatrix}$ using Gram-Schmidt process to ensure orthogonality to the first.
We opt for these four
kernels as bases owing to their simplicity and comprehensiveness. 
This combination covers all directions: horizontal and vertical in the spatial domain, and forward and backward in the temporal domain. Although other alternatives exist, no significant differences have been observed.

\subsection{\network{}}\label{sec:network}

We present the overview of our entire model in Figure~\ref{fig:network}. The overall architecture of our network can be viewed as the general transformation $\mathcal{L}$ in Eq~\ref{eq:formulation}.
There are $N+1$ paths.
The first path, $\mathcal{H}^0$, is implemented as an identity operation; while not strictly a high-frequency extraction, we abuse the notation $\mathcal{H}$ for simplicity. 
The subsequent $N$ extraction modules use the adaptive \module{}, $\{\mathcal{H}^n\}_{n=1}^N$,  each paired with preprocessing modules $\{\mathcal{P}^n\}_{n=1}^N$. 
While these paths employ the same architectural design, they operate with unique parameters, resulting in diverse generated kernels $k_t$. 
Each extraction module takes a blurry video, $X$, and converts it into a unique feature set, 
$\{\mathcal{H}^n(x_t)\}_{n=1}^N$
for the $t$-th frame $x_t$.
This output is then passed to a respective preprocessing module, $\mathcal{P}^n$, to be combined with other features for further analysis.
The result of the $n$-th path for the $t$-th frame is:
\begin{align}
p_t^n = \mathcal{P}^n(h_t) =\mathcal{P}^n(\mathcal{H}^n(x_t)).\label{eq:representation_extractor}
\end{align}

The recurrent extraction module $\mathcal{R}$ takes the feature set $\{p_t^n\}_{n=0}^N$ and the feature from the previous frame, denoted as $h_\text{prev}$, as input. We fuse them using simple concatenation; preliminary experiments showed that more sophisticated fusion structures did not improve 
performance. The warping module $\mathcal{W}$ warps $h_\text{prev}$ with the optical flow computed between $x_\text{prev}$ and $x$ for feature alignment~\cite{chan2021basicvsr}. As a result, the process of the deblurring module can be formulated as:
\begin{align}
    r_t = \mathcal{R}(\{p_t^n\}_{n=0}^N, \mathcal{W}(r_\text{prev}, x_\text{prev}, x_t)),\label{eq:age_hat_rt}
\end{align}
Our model is bidirectional.  $r_\text{prev}$ denotes $r_{t-1}$ and $r_{t+1}$ during forward and backward pass, respectively. In Figure~\ref{fig:network}, 
we make the process for other frames more transparent to highlight the main processing.
Finally, the upsample module $\mathcal{U}$ takes the $r_t$ from the forward ($r_t^f$) and backward passes ($r_t^b$) as input to reconstruct the final output:
\begin{align}
    \hat{y}_t = x_t + \mathcal{U}(r_t^{\text{f}}, r_t^{\text{b}})
\end{align}
The predicted residual $\mathcal{U}(r_t^{\text{f}}, r_t^{\text{b}})$ is equivalent to 
$\mathcal{L}(x_t, \mathcal{H}(\{x_i\}_{i=t-l}^{t+l}))$ in Eq~\ref{eq:formulation}.

The high-frequency extraction modules are integrated into the head stage of the video deblurring model, with the recurrent extraction module $\mathcal{R}$ serving as the primary deblurring component. 
We provide a detailed architecture in Section~\ref{sec:exp_setting}.

Our model strikes a balance between performance and computational budgets in both training and inference. Unlike other works using complicated modules such as transformers~\cite{lin2022flow} and deformable convolutions~\cite{wang2019edvr,jiang2022erdn}, we mainly use a limited number of standard convolutions to build our network. 
This ensures the efficiency. 
Despite its simplicity, this design has been proven effective in improving performance (see Section~\ref{sec:hf_kernel_operation} and \ref{sec:choice_building_kernels}).

\section{Experiments}

\subsection{Experimental Setting}\label{sec:exp_setting}
\noindent \textbf{Datasets.} 
We used the DVD~\cite{su2017deep} and GOPRO datasets~\cite{nah2017deep}. The DVD dataset comprises 61 training and 10 testing videos with $6708$ blurry-sharp image pairs. The GOPRO dataset has 22 training and 11 testing sequences, with $3214$ image pairs. 
Following ~\cite{pan2020cascaded}, we used the version without gamma correction. Each frame is 
of size $1280\times 720$.

\noindent \textbf{Evaluation metrics.} 
The standard metrics for deblurring are the peak signal-to-noise (PSNR) and SSIM~\cite{wang2004image}.
The memory footprint for training each model is measured as the GPU memory required to process a batch of eight $256\times 256$ frames, including both forward and backward propagation.
.
The computational expense of each model is indicated by 
run-time
and number of giga multiply-accumulate operations (GMACs) for the original resolution input.
Lower run-times and GMACs denote greater efficiency.

\noindent \textbf{Training Details.} 
For training, we used a Charbonnier loss~\cite{charbonnier1994two} and the ADAM optimizer~\cite{kingma2014adam} with default hyperparameters, \ie $\beta_1=0.9$ and $\beta_2=0.999$. 
Charbonnier loss preserves edges better than L2 by penalizing outliers less aggressively and avoids artifacts from L1's non-differentiability at zero. The initial learning rate was set to $2\times 10^{-4}$ and decayed with a cosine restart strategy~\cite{loshchilov2016sgdr} with a minimum learning rate of $1\times 10^{-7}$; the number of epochs between two warm starts was set to $300$k. 
Our model is trained for $600$k iterations.
We applied random rotations and flipping for data augmentation, using a batch size of $8$, and $10-$frame training sequences with 
training patches of $256\times256$.  The training used 2 RTX A5000 GPUs.

\noindent \textbf{Architecture Details.} 
The preprocessing module $\mathcal{P}$ is a one-layer convolution followed by two residual dense blocks that downsample the input by a scale factor of 0.25. 
The extraction module $\mathcal{R}$ consists of a single convolution layer followed by 30 residual blocks without batch normalization~\cite{lim2017enhanced}. The output of the residual blocks is upsampled using pixel shuffle modules~\cite{shi2016real} to obtain the residual in $\mathcal{U}$. SPyNet~\cite{ranjan2017optical} is used as the warping module $\mathcal{W}$.

\subsection{Comparison With State-of-the-Art}

\begin{figure}[ht!]
    \centering
    \begin{subfigure}[t]{0.49\linewidth}
        \includegraphics[width=\linewidth]{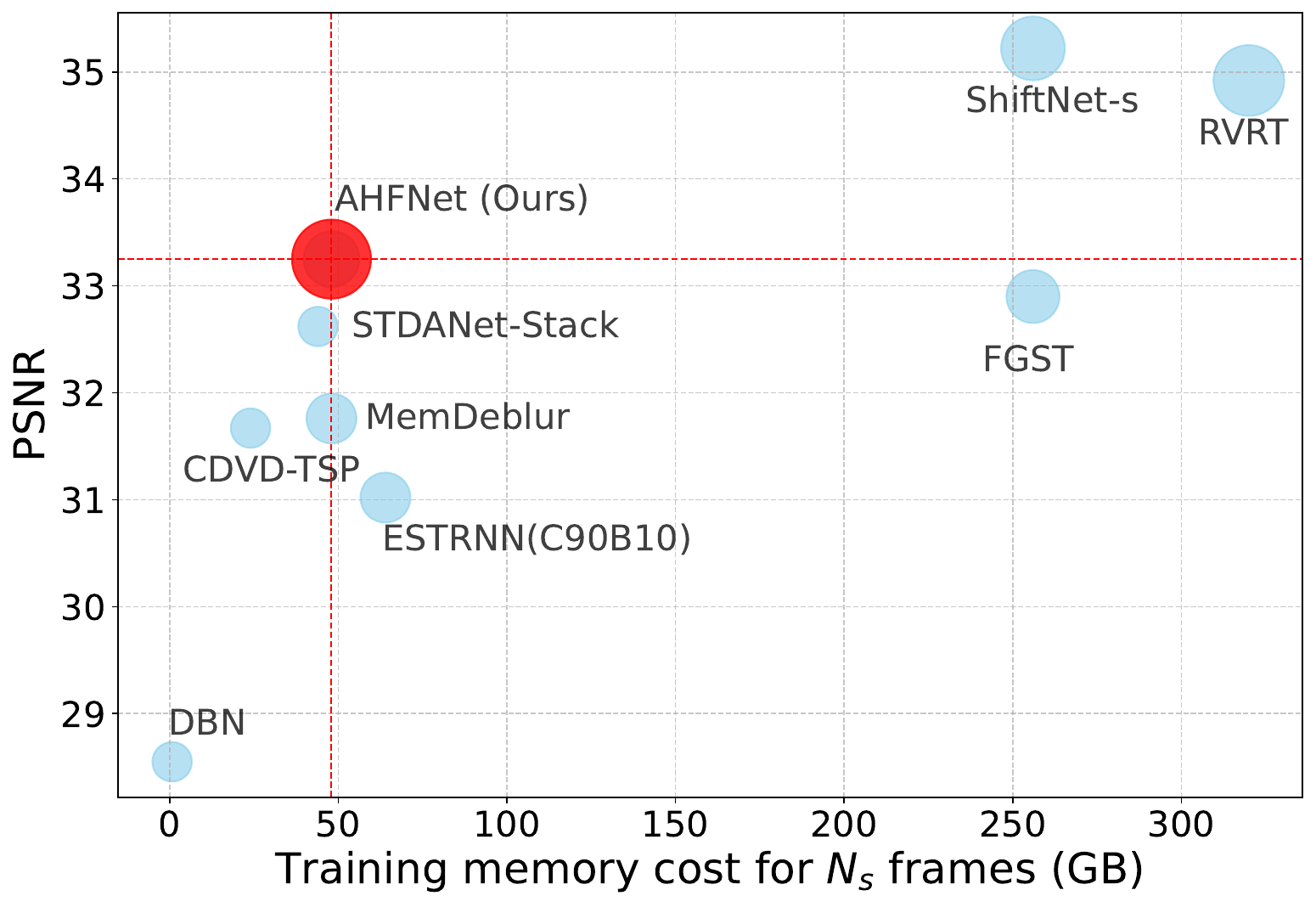}
        \subcaption{$N_s$ frame per sequence \label{fig:tradeoff_psnr_training_cost_Ns}}
    \end{subfigure}
    \begin{subfigure}[t]{0.49\linewidth}
        \includegraphics[width=\linewidth]{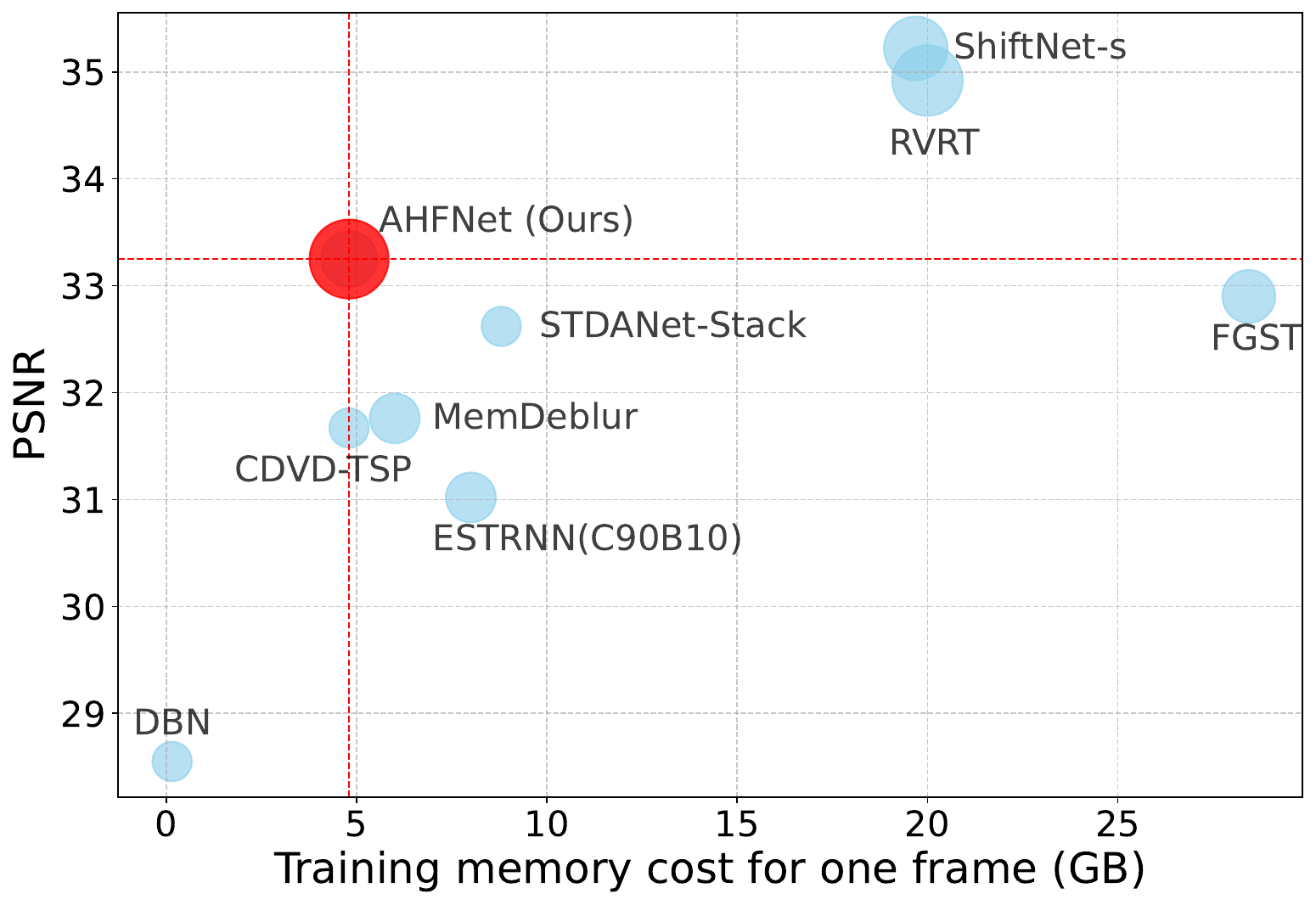}
        \subcaption{$1$ frame per sequence\label{fig:tradeoff_psnr_training_cost_one}}
    \end{subfigure}
    \vspace{-0.1cm}
    \caption{Trade-off between PSNR on GOPRO and the training memory cost. The cost is measured using a $256\times 256$ patch size and a batch size of 8 for processing $N_s$ frames (a) and a single frame (b). The dot size represents $N_s$, which is the number of frames per training sequence used in the official implementation.\label{fig:tradeoff_psnr_training_cost}}
    \vspace{-0.3cm}
\end{figure}

We first present a comparison with all state-of-the-art methods based on the PSNR and training memory cost. Figure~\ref{fig:tradeoff_psnr_training_cost} shows that our model achieves an optimal trade-off between training memory cost and PSNR. 
Using only about $50$GB of memory for training one batch, we achieve a PSNR of 33.25, outperforming many models, including FGST~\cite{lin2022flow}, which requires more than $250$GB, as shown in Figure~\ref{fig:tradeoff_psnr_training_cost_Ns}.
While models like ShiftNet~\cite{li2023simple} and RVRT~\cite{liang2022recurrent} achieve higher PSNRs, they need at least $5$ times the GPU memory for training. The left image shows the full GPU memory their official implementations require, which is about $250$GB (8 GPUs). 
We further provide Figure~\ref{fig:tradeoff_psnr_training_cost_one} to show the memory cost when processing a single frame during training, as different models uses different number of frames per training sequence. For example, our method uses 10 frames per training sequence, while RVRT uses 16. When fixing this variable factor to be the same, our model still locates at the top-left position in Figure~\ref{fig:tradeoff_psnr_training_cost_one}, indicating high performance with fewer training resources.
One reason for the high training cost of other models is their extensive use of complex operations, such as self-attention, to continually refine specific features.

\begin{figure}
    \centering
    \includegraphics[width=0.75\linewidth]{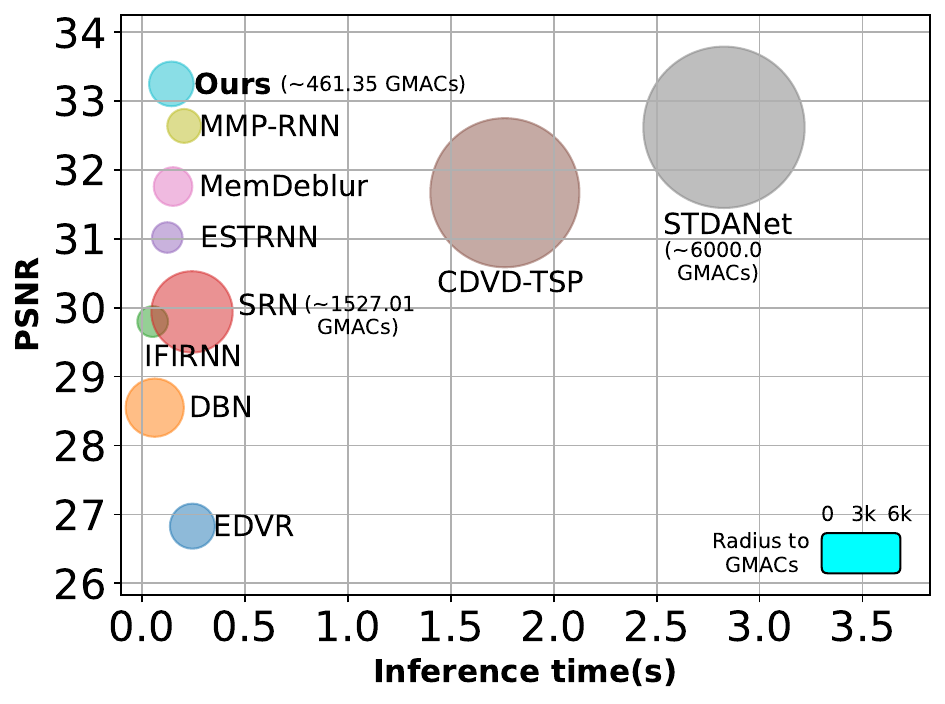}
    \caption{Inference time for a single $1280\times 720$ frame 
    vs. PSNR on the GOPRO dataset~\cite{nah2017deep}.  The dot size represents GMACs.
    Compared models share a similar training memory footprint.  
    Our method achieves the best PSNR with minimal inference time.}
    \label{fig:teaser_photo}
\end{figure}

To ensure a fair comparison and thorough comparison, we evaluate our model against state-of-the-art models trained with similar training budgets on the GOPRO~\cite{nah2017deep} and DVD~\cite{su2017deep} datasets, as presented in 
Tables~\ref{table:sota_gopro} and \ref{table:sota_dvd}, respectively.  
Hence, ShiftNet, RVRT and FGST are not included.
In our presentation of results, we detail the complexity metrics only in Table~\ref{table:sota_gopro} and omit these details from Table~\ref{table:sota_dvd}. 
Our model achieves state-of-the-art with relatively fewer GMACs and less run-time than models with similar training budgets. 
Compared with MMP-RNN, our PSNR is $0.61$dB higher with 70\% of the run-time despite larger GMACs. 
{MMP-RNN’s main building block is the residual dense block. The dense connectivity in MMP-RNN leads to more memory operations and is less optimized than normal one, which decreases its runtime. }
Compared with ERDN and STDANet, our model is superior in performance and efficiency.
Figure~\ref{fig:teaser_photo} illustrates the relationship between PSNR and inference time, with the radius of each model denoting GMACs. This visually demonstrates the efficiency and effectiveness of our proposal.
{Qualitatively, Fig.~\ref{fig:qualitative_gopro} shows that our model excels in producing the sharpest curves of the car 
, which is originally severely blurred.}

\begin{table}
    \caption{Performance comparison to models with a similar training memory footprint on the GOPRO dataset
    \label{table:sota_gopro}}
    \vspace{-0.3cm}
    \centering
    \resizebox{\linewidth}{!}{
    \def\arraystretch{1.1}
    \begin{tabular}{@{\extracolsep{4pt}}cccccc@{}}
        \Xhline{3\arrayrulewidth}
        Model & PSNR & SSIM & GMACs & Params (M) & Time(s) \\
        \hline
        DBN~\cite{su2017deep} & 28.55 & 0.8595 & 784.75 &15.31 & 0.063\\
        IFIRNN ($\text{c}_2\text{h}_3$)~\cite{nah2019recurrent} & 29.80 & 0.8900 & 217.89 & 1.64 & 0.053 \\
        SRN~\cite{tao2018scale} & 29.94 & 0.8953 & 1527.01 &10.25 & 0.244 \\
        ESTRNN ($\text{C}_{90}\text{B}_{10}$)~\cite{zhong2020efficient} & 31.02 & 0.9109 & 215.26 & 2.47 & 0.124 \\
        CDVD-TSP~\cite{pan2020cascaded} & 31.67 & 0.9279 & 5122.25 & 16.19 & 1.763 \\
        MemDeblur~\cite{ji2022multi} &  31.76 &  0.9230 & 344.49  & 6.99 & 0.152\\
        ERDN~\cite{jiang2022erdn} & 32.48 & 0.9329 & 29944.57 & 45.68 & 5.094 \\
        STDANet-Stack~\cite{zhang2022spatio} & 32.62 & 0.9375 & 6000.00 & 13.84 & 2.827 \\
        MMP-RNN~\cite{wang2022efficient} & 32.64 & 0.9359 & 264.52 & 4.05 & 0.206 \\
        \hline
        \network & 33.25 & 0.9439 & 461.35 & 6.75 & 0.144 \\
        \Xhline{3\arrayrulewidth}
    \end{tabular}
    }
\end{table}

\begin{table}
    \caption{
    Performance comparison to models with a similar training memory footprint on the DVD dataset. 
    As ARVo~\cite{li2021arvo} did not provide details such as MACs and there is no official code available, these values are marked as ``-". \label{table:sota_dvd}}
    \vspace{-0.3cm}
    \centering
    \def\arraystretch{1.1}
    \resizebox{\linewidth}{!}{
    \begin{tabular}{@{\extracolsep{4pt}}cccccc@{}}
        \Xhline{3\arrayrulewidth}
        Model & PSNR & SSIM & GMACs & Params (M) & Time(s) \\
        \hline
        SRN & 30.53 & 0.8940 & 1527.01 & 10.25 & 0.244\\
        IFIRNN ($\text{c}_2\text{h}_3$) & 30.80 & 0.8991 & 217.89 & 1.64 & 0.053 \\
        EDVR & 31.82 & 0.9160 & 468.25 & 20.04 & 0.246\\
        CDVD-TSP & 32.13 & 0.9268 & 5122.25 & 16.19 & 1.763\\
        ARVo~\cite{li2021arvo} & 32.80 & 0.9352  &  -&  - & - \\
        STDANet-Stack & 33.05 &  0.9374 & 6000.00 & 13.84  & 2.827 \\
        \hline
        \network & 33.19 & 0.9400 & 461.35 &  6.75 & 0.144\\
        \Xhline{3\arrayrulewidth}
    \end{tabular}
    }
    \vspace{-0.5cm}
\end{table}

\begin{figure*}[!t]
	\captionsetup{font=small}
	
	\centering
	\scriptsize
	
	\renewcommand{\h}{0.105}
	\renewcommand{\wa}{0.12}
	\newcommand{\wb}{0.16}
	\renewcommand{\g}{-0.7mm}
	\renewcommand{\tabcolsep}{1.8pt}
	\renewcommand{\arraystretch}{1}
	\resizebox{1.00\linewidth}{!} {
		\begin{tabular}{cc}
			
			\renewcommand{\name}{figures/sota_visualization/gopro_1/}
			\renewcommand{\namepatch}{figures/sota_visualization/gopro_1/patch2_}
			\renewcommand{\h}{0.12}
			\renewcommand{\w}{0.2}
			\begin{tabular}{cc}
				\begin{adjustbox}{valign=t}
					\begin{tabular}{c}%
		         	\includegraphics[width=0.31\textwidth]{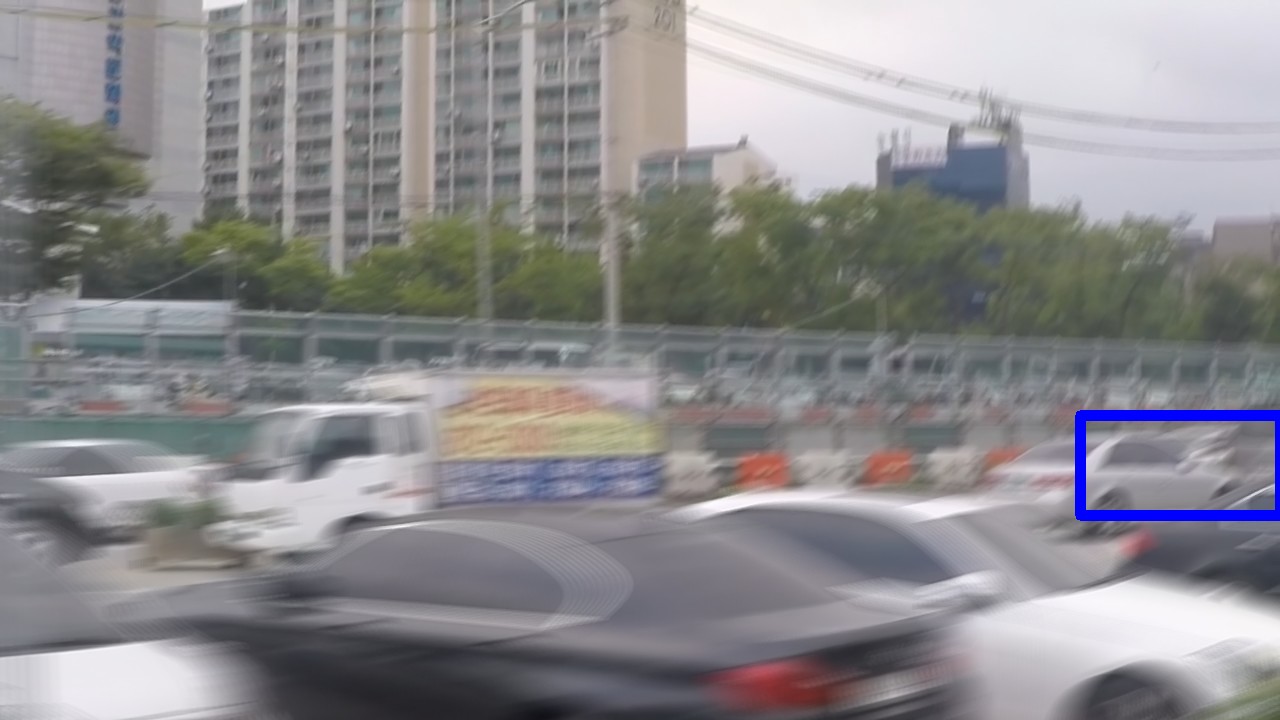}
						\\
					 Input blurry frame on GOPRO~\cite{nah2017deep} 
					\end{tabular}
				\end{adjustbox}
				\begin{adjustbox}{valign=t}
					\begin{tabular}{cccccc}
						\includegraphics[width=\w \textwidth]{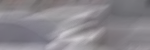} \hspace{\g} &
						\includegraphics[width=\w \textwidth]{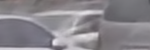} \hspace{\g} &
						\includegraphics[width=\w \textwidth]{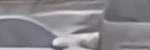} \\
						Input &
                        ESTRNN~\cite{zhong2020efficient} &
						CDVD-TSP~\cite{pan2020cascaded} 
						\\
						\\
						\includegraphics[width=\w \textwidth]{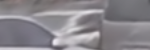} &
                        \includegraphics[width=\w \textwidth]{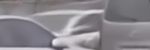} \hspace{\g} &
						\includegraphics[width=\w \textwidth]{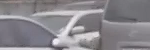} \hspace{\g} \\
                        MMP-RNN~\cite{wang2022efficient} &
						Ours \hspace{\g} &
						GT 
						\\
					\end{tabular}
				\end{adjustbox}
			\end{tabular}
		\end{tabular}
	}

        \resizebox{1.00\linewidth}{!} {
		\begin{tabular}{cc}
			
			\renewcommand{\name}{figures/sota_visualization/dvd_2/}
			\renewcommand{\namepatch}{figures/sota_visualization/dvd_2/patch3_}
			\renewcommand{\h}{0.12}
			\renewcommand{\w}{0.2}
			\begin{tabular}{cc}
				\begin{adjustbox}{valign=t}
					\begin{tabular}{c}%
		         	\includegraphics[width=0.31\textwidth]{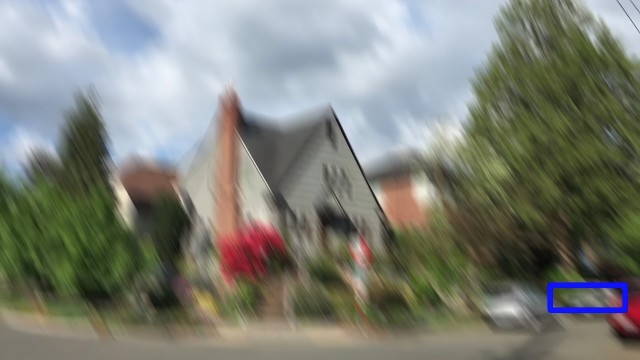}
						\\
					 Input blurry frame on DVD~\cite{su2017deep} 
					\end{tabular}
				\end{adjustbox}
				\begin{adjustbox}{valign=t}
					\begin{tabular}{cccccc}
						\includegraphics[width=\w \textwidth]{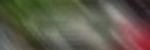} \hspace{\g} &
						\includegraphics[width=\w \textwidth]{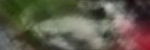} \hspace{\g} &
						\includegraphics[width=\w \textwidth]{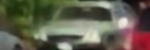} \\
						Input &
                            DBN~\cite{su2017deep} &
						CDVD-TSP~\cite{pan2020cascaded} 
						\\
						\\
						\includegraphics[width=\w \textwidth]{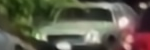} &
                        \includegraphics[width=\w \textwidth]{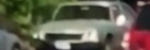} \hspace{\g} &
						\includegraphics[width=\w \textwidth]{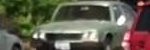} \hspace{\g} \\
						
                            STDAN~\cite{zhang2022spatio} &
						Ours \hspace{\g} &
						GT 
						\\
					\end{tabular}
				\end{adjustbox}
			\end{tabular}
		\end{tabular}
	}
 
        \vspace{-0.3cm}
	\caption{Qualitative comparisons to models with a similar training memory footprint.}
	\label{fig:qualitative_gopro}
 \vspace{-0.5cm}
\end{figure*}

\subsection{Predefined versus Learned Kernels}\label{sec:hf_kernel_operation}
In this section, we evaluate the dynamic approach to generating kernels. 
There are two primary methods for dynamically generating convolutional kernels $k_t$: (1) directly predicting $k_t$~\cite{jia2016dynamic}, and (2) predicting the coefficients $\{\alpha_{t,j}\}_{j=1}^M$ and using basis kernels $\{\tilde{k}_j\}_{j=1}^M$ to construct $k_t$~\cite{xia2020basis}. 
As mentioned in Section~\ref{sec:approach}, our method adopts the second generation method because we aim to constrain the generated kernel $k_t$ to be high-pass, which the first generation method does not ensure. 
For each generation strategy, we create a variant for comparison. 
Specifically, the first strategy yields the variant denoted as “3D Conv,” and the second yields “Naive Basis Kernels”, where $\{\tilde{k}_j\}_{j=1}^M$ consists of standard basis with patterns like $[1 \ 0 \ 0 \ \dots \ 0]$, $[0 \ 1 \ 0 \ \dots \ 0]$, ..., $[0 \ 0 \ 0 \ \dots \ 1]$. 
We opt for these specific kernels because they represent standard bases in the vector space. 
It is obvious that substituting these basis kernels with high-pass kernels yields our method. 
Both the “3D Conv” and “Naive Basis Kernels” variants have similar complexity to our approach but are not specifically designed to extract high frequencies. Each variant was trained for 300k iterations.

Table~\ref{table:abl_3dconv} shows that our proposed module achieves the highest PSNR and SSIM, demonstrating improvement without adding complexity. 
The PSNR difference between the two generation strategies is 0.02 dB, but the direct generation method requires predicting more learnable parameters since the entire kernel is of size $T_k\! \times\! H_k \!\times\! W_k$. In contrast, the coefficient generation strategy is more efficient and deals with only $M$ learnable parameters.

Moreover, to study the effects of learning in different frequency subbands, we perform the DFT on the output images, normalize their frequencies to the range $[0, 2\pi]$, and divide the Fourier spectra of the outputs from each variant into 10 equal subbands. 
Each subband covers an interval of $2\pi/10$, from $[0, 2\pi/10]$ to $(18\pi/10, 2\pi]$. 
We then calculate the relative average mean squared error (MSE) for each subband on the GOPRO dataset. The MSE values are relative to the variant without the dynamic module, which can result in negative values.

\begin{figure}
        \centering
        \includegraphics[width=0.75\linewidth]{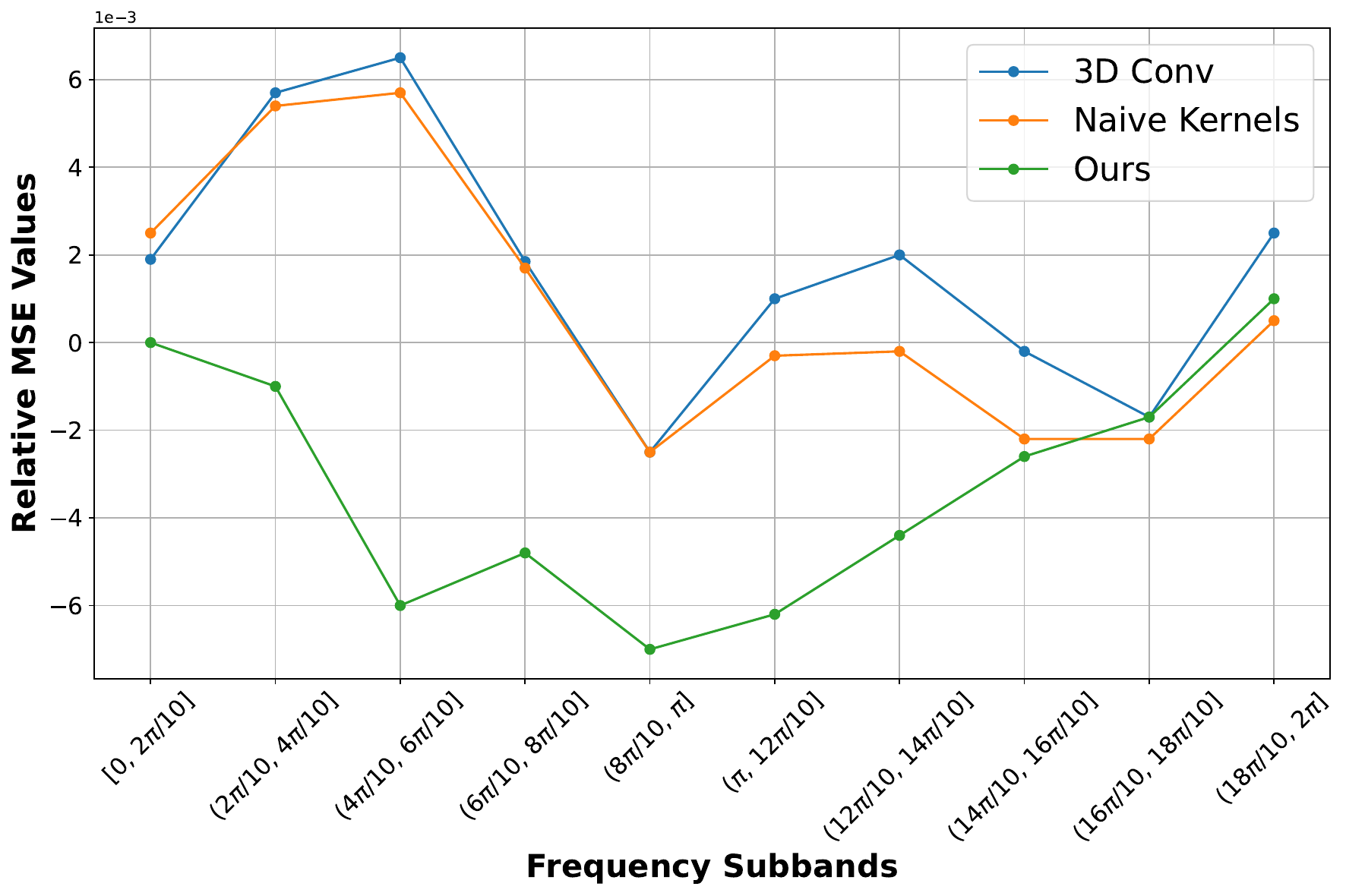}
        \vspace{-0.2cm}
        \caption{Comparison of learning in various frequency subbands
        . Lower MSE values indicate better performance.\label{fig:frequency_learning}}
        \vspace{-0.2cm}
\end{figure}

Figure~\ref{fig:frequency_learning} presents the result, where a lower curve indicates better performance. 
Our method's curve, depicted in green, consistently shows the lowest values across most subbands, indicating superior performance. Despite slight reductions in performance at the highest frequency areas, likely due to training variance, our approach remains notably effective across most of the subbands.

Additionally, we assess scenarios with varying numbers of paths for extraction, \ie $N$. The result is in Table~\ref{table:abl_N}.
Increasing the number of paths can further enhance performance, but this comes at the expense of increased computational resources.

\begin{table}
    \caption{ 
    Validation of the HF extraction module on GOPRO.\label{table:abl_3dconv}}
    \vspace{-0.2cm}
    \centering
    \resizebox{\linewidth}{!}{
    \def\arraystretch{1.1}
    \begin{tabular}{@{\extracolsep{4pt}}ccccc@{}}
        \Xhline{2\arrayrulewidth}
        &  PSNR & SSIM & GMACs & Params (M) \\
        \hline
        3DConv & 32.82 & 0.9384 & 462.71 & 6.75 \\ 
        Naive Kernels & 32.80 & 0.9378 & 461.35 & 6.75\\
        \network{} & \textbf{32.89} & \textbf{0.9398} & 461.35 & 6.75 \\
        \Xhline{2\arrayrulewidth}
    \end{tabular}
    }
    \vspace{-0.5cm}
\end{table}

\begin{table}
    \caption{Ablation on HF extraction module count $N$ on GOPRO.\label{table:abl_N}}
    \vspace{-0.2cm}
    \centering
    \def\arraystretch{1.1}
    \begin{tabular}{@{\extracolsep{4pt}}ccccc@{}}
        \Xhline{2\arrayrulewidth}
        $N$ & 0 & 2 & 4 & 6\\
        \hline
        PSNR & 32.56  & 32.68& 32.71 & \textbf{32.89}\\
        SSIM &  0.9366 &0.9372 &0.9371 & \textbf{0.9398} \\
        GMACs & 410.73 & 427.66 &  444.51 &461.35 \\
        \Xhline{2\arrayrulewidth}
    \end{tabular}
\end{table}

\subsection{Choice of 
basis kernels
}\label{sec:choice_building_kernels}

\begin{table}
    \centering
    \caption{Experiments on various high-pass filters on GOPRO.}\label{table:abl_M}
    \vspace{-0.2cm}
    \def\arraystretch{1.1}
    \begin{tabular}{@{\extracolsep{4pt}}cccc@{}}
        \Xhline{3\arrayrulewidth}
        Added Representation & PSNR & SSIM & GMACs\\
        \hline
        + RGB$\times$1 & 32.47 & 0.9363 & 400.82\\
        + $\nabla^2 x$ & 32.37 & 0.9335 & 402.26\\
        + $\nabla_t x$ & 32.41 & 0.9340 & 400.82\\
        + $\nabla x$ (Kirsch) & 32.51 & 0.9356 & 403.76 \\
        + $\nabla x$ (Sobel) & 32.53 & 0.9363 & 403.76 \\
        \hline
        + RGB$\times$2 & 32.45 & 0.9356 & 414.39 \\
        + $\nabla x$ (Sobel) + $\nabla_t x$ & \textbf{32.78} & \textbf{0.9394} & 417.28\\
        \hline
        + Random& 31.98 & 0.9312 & 444.51\\
        \Xhline{3\arrayrulewidth}
    \end{tabular}
    \vspace{-0.4cm}
\end{table}

\begin{figure*}[!t]
	\captionsetup{font=small}
	
	\centering
	\scriptsize
	
	\renewcommand{\h}{0.105}
	\renewcommand{\wa}{0.12}
	\newcommand{\wb}{0.16}
	\renewcommand{\g}{-0.7mm}
	\renewcommand{\tabcolsep}{1.8pt}
	\renewcommand{\arraystretch}{1}
	\resizebox{1.00\linewidth}{!} {
		\begin{tabular}{cc}
			
		      \renewcommand{\name}{figures/learned_vis/}
			\renewcommand{\h}{0.12}
			\renewcommand{\w}{0.2}
			\begin{tabular}{cc}
				\begin{adjustbox}{valign=t}
					\begin{tabular}{c}%
		         	\includegraphics[width=0.31\textwidth]{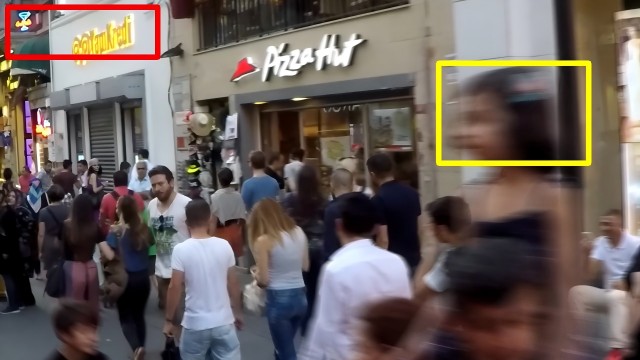}
						\\
					 Input blurry frame~\cite{nah2017deep} 
					\end{tabular}
				\end{adjustbox}
				\begin{adjustbox}{valign=t}
					\begin{tabular}{cccccc}
						\includegraphics[width=\w \textwidth]{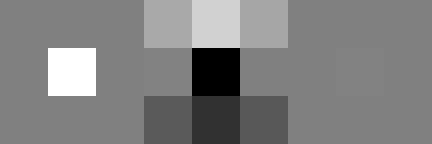} \hspace{\g} &
						\includegraphics[width=\w \textwidth]{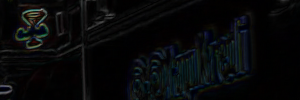} \hspace{\g} &
						\includegraphics[width=\w \textwidth]{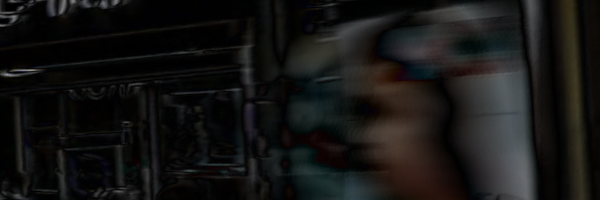} \\
						$k^1$ &
						Red region  & 	
                            Yellow region
						\\
                            \\
						\includegraphics[width=\w \textwidth]{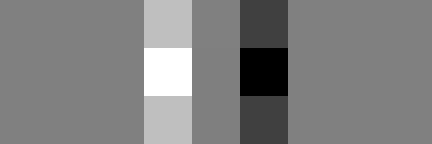} \hspace{\g} &
						\includegraphics[width=\w \textwidth]{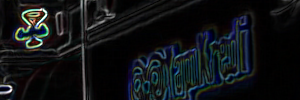} \hspace{\g} &
						\includegraphics[width=\w \textwidth]{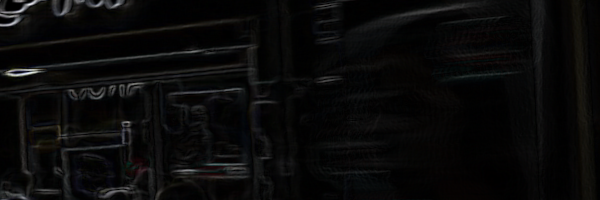} \\
						$k^4$ &
						Red region  & 	
                            Yellow region
						\\
                            \\
					\end{tabular}
				\end{adjustbox}
			\end{tabular}
		\end{tabular}
	}

 \vspace{-0.5cm}
	\caption{Examples of learned kernels and features on GOPRO dataset (please zoom in for a better view).}
	\label{fig:learned_vis}
 \vspace{-0.5cm}
\end{figure*}

We evaluate our choice of basis kernels $\{\tilde{k}_j\}_{j=1}^M$ (see Sec.~\ref{sec:module}) by comparing with two other options. First, we use standard high-pass filters, including first- ($\nabla x$, \eg Sobel and Kirsch) and second-order kernels ($\nabla^2 x$, \eg Laplacian), and a temporal gradient filter ($\nabla_t x$) given by $[[1/2], [-1], [1/2]]$. Secondly, we follow ~\cite{wang2022anti} to randomly generate high-pass filters as our
basis kernels. 
Specifically, given 
an arbitrary matrix $\mathbf{A}$
, the corresponding high-pass filter is $(\mathbf{I} - \frac{1}{n}\mathbf{1}\mathbf{1}^T )\mathbf{A}$, where $\mathbf{I}$ is an identity matrix, $n$ is the number of elements in $\mathbf{A}$ and $\mathbf{1}$ is a vector filled with ones.
This variant is referred to as ``+ Random'', with the 
basis kernels
randomly generated for each training run.
We keep these kernels relatively simple, as the subsequent preprocessing module $\mathcal{P}$ (see Fig.~\ref{fig:network}) 
use non-linearity to facilitates hierarchical feature extraction, evolving low-level features into complex representations
We used the same architecture in Section~\ref{sec:network} and replaced these choices with the \module{}.

Comparison results are reported in Table~\ref{table:abl_M}.  As a control study to match the complexity, we also show variants ``+ RGB$\times k$'', which repeats the early module $(\mathcal{P}\!\circ\! \mathcal{H})$ $k$ times.
Adding first-order filters improves PSNR by up to $0.06$dB across variants, while second-order and temporal gradients do not boost performance. Second-order spatial gradients are more noise-sensitive. 
{Using only the temporal gradient 
results in inaccurate motion estimations because consecutive frames can have different levels of blur.}
Integrating first-order and temporal filters (``+ $\nabla x$ (Sobel) + $\nabla_t x$") 
significantly enhances image quality, increasing PSNR by 0.33dB and SSIM by 0.0038 compared to the baseline
(``+ RGB $\times$ 2").  
{
The first-order feature often positively correlates with blur characteristics and helps refine temporal feature. Therefore, their combination achieves the best results.}
Moreover, compared to the ``+ Random'' variant, our method elevates PSNR by 0.8dB and SSIM by 0.0082, 
with fewer GMACs. 
The ``+ Random" variant has higher computational costs, as it requires more efforts to generate high-pass kernels. 
This indicates that predefined high-pass filters, especially the combination of Sobel and temporal gradients, outperform randomly generated ones. These findings guide our choice of kernels for the \network{} architecture.

\subsection{Visualizations}
Figure~\ref{fig:learned_vis} visualizes the learned kernels, $\{k_t^n\}_{n=1}^N$, and the high-frequency features extracted by our learned kernels, $\{h_t^n\}_{n=1}^N$. 
We select two out of $N=6$ representations for visualization.
Kernels are normalized to $[0, 1]$, where white, gray, and black signify values of 1, 0.5, and 0, respectively.
Additional visualizations of kernels and representations are provided in the supplementary material.

The first row combines both spatial and temporal details. The second row focuses exclusively on the current frame, resulting in a spatial gradient map, where shop signs are clearly visible, but human faces are barely distinguishable.
These two features highlight different aspects of the same frame, showing the effectiveness of our model in extracting varied features, which enhances the deblurring process.
{This diversity arises because each high-pass kernel basis captures high frequencies from different directions, with coefficients predicted by modules initialized with varied parameters. Due to the non-convex nature of neural networks, even slight differences in initialization lead to distinct parameter updates, further diversifying the features.}
Our choice of basis kernel is currently simple, resulting in similarly simple generated kernels. However, more complex basis kernels may be advantageous.

\section{Conclusion}
We emphasize the crucial role of high frequencies in video deblurring and formulate our algorithm to explicitly extract and utilize these frequencies.
Experimental results underscore the significance of manual high-frequency extraction. Notably, the combination of first-order and temporal gradients substantially enhances performance.

Progressing further, we introduce adaptability in high-frequency extraction by generating the coefficients of several high-pass filter kernels. 
Initially, we show that the linear combination of positive coefficients and high-pass filter kernels continues to function as a high-pass filter.
This assurance confirms our generated kernel's capacity to extract high frequencies from the input feature efficiently. By integrating this module into our model, we attain state-of-the-art results in video deblurring datasets regarding performance and complexity among low-budget models. 
Visual demonstrations further highlight the effectiveness of our proposed approach. Our approach shows enhanced performance across various frequency subbands and improved filtered results in the spatial domain.

{\small
\bibliographystyle{ieee_fullname}
\bibliography{main}
}

\end{document}